%% file: eacl2021.tex
\documentclass[11pt,a4paper]{article}
\usepackage[hyperref]{eacl2021}
\usepackage{times}
\usepackage{latexsym}
\usepackage{amsmath,amssymb,graphicx,wasysym,amssymb}
\usepackage{mathtools}
\usepackage{multirow,booktabs,xcolor,colortbl}

\usepackage{microtype}
\usepackage{todonotes}
\usepackage{enumitem}
\usepackage{subfig}
\usepackage{xcolor}

\aclfinalcopy 

\newcommand{\red}[1]{\textcolor{black}{#1}}
\def\ie{i.e.\ }

\newcommand{\B}{\textbf}
\newcommand{\I}{\textit}
\newcommand{\T}{\texttt}

\usepackage{xspace}

\newcommand{\init}{\texttt{RL-init}\xspace}
\newcommand{\att}{\texttt{RL-att}\xspace}
\newcommand{\env}{\texttt{RL-env}\xspace}
\newcommand{\base}{\texttt{RL-base}\xspace}
\newcommand{\initatt}{\texttt{RL-init-att}\xspace}
\newcommand{\envinitatt}{\texttt{RL-env-init-att}\xspace}

\title{Exploiting Multimodal Reinforcement Learning for Simultaneous Machine Translation}

\author{Julia Ive$^{1}${\normalfont,} Andy Mingren Li$^1${\normalfont,} Yishu Miao$^1${\normalfont,} Ozan Caglayan$^{1}${\normalfont,} \\[.2em] {\bf Pranava Madhyastha$^1$ \and Lucia Specia$^{1,2,3}$}\\[.3em]
Imperial College London$^1$,\, University of Sheffield$^2$,\, ADAPT - Dublin City University$^3$\\
         \texttt{\small j.ive@ic.ac.uk,
         andy.li16@imperial.ac.uk,
         y.miao20@imperial.ac.uk,
         o.caglayan@ic.ac.uk} \\
         \texttt{\small l.barrault@sheffield.ac.uk, l.specia@ic.ac.uk}\\
}

\date{}

\begin{document}
\maketitle
\begin{abstract}
This paper addresses the problem of simultaneous machine translation (SiMT) by exploring two main concepts: (a) adaptive policies to learn a good trade-off between high translation quality and low latency; and (b) visual information to support this process by providing additional (visual) contextual information which may be available before the textual input is produced. For that, we propose a multimodal approach to simultaneous machine translation using reinforcement learning, with strategies to integrate visual and textual information in both the agent and the environment. We provide an exploration on how different types of visual information and integration strategies affect the quality and latency of simultaneous translation models, and demonstrate that visual cues lead to higher quality while keeping the latency low. 
\end{abstract}

\section{Introduction}
\label{sec:intro}

\input{intro.tex}

\section{Related Work}
\label{sec:relwork}

\input{relwork.tex}

\section{Methods}
\label{sec:method}
\input{method.tex}

\section{Experimental Setup}
\label{sec:setup}
\input{setup.tex}

\section{Results}
\label{sec:results}
\input{results.tex}

\section{Conclusion}
\label{sec:conclusion}
\input{conclusion.tex}

\section*{Acknowledgments}
The authors thank the anonymous reviewers for their useful feedback. This work was supported by the MultiMT (H2020 ERC Starting Grant No. 678017) project. The work was also supported by the Air Force Office of Scientific Research (under award number FA8655-20-1-7006) project. Andy Mingren Li was supported by the Imperial College London UROP grant.

\bibliography{anthology,eacl2021}
\bibliographystyle{acl_natbib}


\end{document}

%% file: intro.tex
Research into automating real-time interpretation has explored deterministic and adaptive approaches to build policies that address the issue of translation delay~\cite{ryu-etal-2006-simultaneous,cho2016can,gu-etal-2017-learning}. In another recent development, the availability of multimodal data (such as visual information) has driven the community towards multimodal approaches for machine translation (MMT)~\cite{specia-etal-2016-shared,elliott-etal-2017-findings,barrault-etal-2018-findings}.
Although deterministic policies have been recently explored for simultaneous MMT~\cite{caglayan2020simultaneous,imankulova-etal-2020-towards}, there are no studies regarding how multimodal information can be exploited to build flexible and adaptive policies for simultaneous machine translation (SiMT).

Applications of reinforcement learning (RL) for unimodal SiMT have highlighted the challenges for the agent to maintain good translation quality while learning an optimal translation path (\ie a sequence of \T{READ/WRITE} decisions at every time step) \cite{grissom-ii-etal-2016-incremental,gu-etal-2017-learning,alinejad-etal-2018-prediction}.  

Incomplete source information will have detrimental effect especially in the cases where significant restructuring is needed while translating from one language to another. 

In addition, the lack of information generally leads to high variance during the training in the RL setup. 
We posit that {\bf multimodality} in adaptive SiMT could help the agent by providing extra signals, which would in turn improve training stability and thus the quality of the estimator and translation decoder.

In this paper, we present the first exploration on multimodal RL approaches for the task of SiMT.
 
As visual signals, we explore both image classification features as well as visual concepts, which provide global image information and explicit object representations, respectively. For RL, we employ the Policy Gradient method with a pre-trained neural machine translation model acting as the environment. 

As the SiMT model is optimised for both translation quality and latency, we apply a combined reward function that consists of a decomposed smoothed BLEU score and a latency score. To integrate visual and textual information, we propose different strategies that operate both on the agent (as prior information or at each step) and the environment side.

In experiments on standard datasets for MMT, our models achieve the highest BLEU scores on most settings without significant loss on average latency, as compared to strong SiMT baselines. A qualitative analysis shows that the agent benefits from the multimodal information by grounding language signals on the images.  

Our {\bf main contributions} are as follows: (1) we propose the first multimodal approach to simultaneous machine translation based on adaptive policies with RL, introducing different strategies to integrate visual and textual information (Sections~\ref{sec:method} and \ref{sec:setup}); (2) we show how different types of visual information and integration strategies affect the quality and latency of the models (Section~\ref{sec:results}); (3) we demonstrate that providing visual cues to both agent and environment is beneficial: models achieve high quality while keeping the latency low (Section~\ref{sec:results}).

%% file: relwork.tex
In this section, we first present background and related work on SiMT, and then discuss recent work in MMT and multimodal RL. 

\subsection{Simultaneous Machine Translation} 

In the context of neural machine translation (NMT), \citet{cho2016can} introduce a greedy decoding framework where simple heuristic waiting criteria are used to decide whether the model should read more source words or instead write a target word. \citet{gu-etal-2017-learning} utilise a pre-trained NMT model in conjunction with an RL agent whose goal is to learn a \T{READ/WRITE} policy by maximising quality and minimising latency. \citet{alinejad-etal-2018-prediction} further extend the latter approach by adding a \T{PREDICT} action with an aim to capture the anticipation of the next source word. 
\citet{ma-etal-2019-stacl} propose an end-to-end, fixed-latency framework called `wait-$k$' which allows \I{prefix-to-prefix} training using a deterministic policy: the agent starts by reading a specified number of source tokens ($k$), followed by alternating \T{WRITE} and \T{READ} actions. 

Other approaches to SiMT include re-translation of previous outputs depending on new outputs~\cite{arivazhagan-et-al-retranslation,niehues2018} or learning adaptive policies guided by a heuristic or alignment-based approaches~\cite{zheng-etal-2019-simpler,arthur-et-al-learning}.
A general theme in these approaches is their reliance on \I{consecutive} NMT models pre-trained on full-sentences. However, \citet{dalvi-etal-2018-incremental} discuss potential mismatches between the training and decoding regimens of these approaches and propose to perform fine-tuning of the models using chunked data or prefix pairs.

\subsection{Multimodal Machine Translation}
MMT aims at improving the quality of automatic translation using additional sources of information~\cite{sulubacak2019multimodal}. Different methods for fusing textual and visual information have been proposed. These include initialising the textual encoder or decoder with the visual information~\cite{elliott-kadar-2017-imagination,caglayan-etal-2017-lium}, combining the visual information through spatial feature maps using soft attention~\cite{caglayan-etal-2016-multimodality,libovicky-helcl-2017-attention,huang-etal-2016-attention,calixto-etal-2017-doubly}, and projecting a summary of the visual representations to a common context space via a trained projection matrix~\cite{calixto-liu-2017-incorporating,caglayan-etal-2017-lium,elliott-kadar-2017-imagination,gronroos-etal-2018-memad}. Further, recent work has also focused on exploring Multimodal Pivots~\cite{hitschler-etal-2016-multimodal} and latent variable models~\cite{calixto-etal-2019-latent} in the context of multimodal machine translation. In this paper, we explore all these strategies, and also the use of {\em visual concepts}, similar to the approach by~\citet{ive-etal-2019-distilling}.

\subsection{Multimodal Reinforcement Learning} 

Previous work has explored RL with language inputs \cite{andreas2017modular,bahdanau2018learning,goyal2019using} by  making use of language to improve the policy or reward function: 
for example, the task of navigating in the world grid environment using language instructions~\cite{andreas2017neural}.

Alternatively, RL with language output can be shaped as  sequential decision making for language generation, while conditioning on other modalities. This includes image captioning~\cite{ren2017deep}, video captioning \cite{wang2018video}, question answering \cite{das2018embodied}, and text-based games \cite{cote2018textworld}. Our study sits somewhere in between these different types of work. We have both the source language and respective images as input and the target language as output. Our agent is focused only on learning the \T{READ} and \T{WRITE} actions while the translation model is fixed for simplicity.

The central aim of the agent is learning to capture the relevant structures and relations of the modalities that can lead to a better SiMT system.

%% file: method.tex
We first present the architectures for consecutive and baseline fixed policy simultaneous MT (Section \ref{sec:baselines}). Then we introduce our RL approaches, both the baseline, the proposed multimodal extension (Section \ref{sec:method_sim_rl}), and the visual features used by all multimodal approaches (Section \ref{sec:visfeatures}). 


\subsection{Baselines}\label{sec:baselines}
\paragraph{Unimodal MT.}
\label{sec:method_nmt}
We implement a standard encoder-decoder baseline with attention~\cite{Bahdanau2014} which incorporates a two-layer encoder and a two-layer decoder with GRU~\cite{cho-etal-2014-learning} units. Given a source sequence of embeddings $X\mathrm{=} \{x_1,\dots,x_S\}$ and a target sequence of embeddings $Y\mathrm{=}\{y_1,\dots,y_T\}$, the encoder first computes the sequence of hidden states $H\mathrm{=}\{h_1,\dots,h_S\}$ unidirectionally. 

The attention layer receives $H$ as \textit{key-values} whereas the hidden states of the first decoder GRU provide the \textit{queries}. The context vector $c_t^{\mathbf{T}}$ produced by the attention layer is given as input to the second GRU. Finally, the output token ($y_t$) probabilities are obtained by applying a softmax layer on top of the concatenation of the previous word embedding, context vector and the second GRU's hidden state.
 
For {\bf consecutive NMT}, all source tokens are observed before the decoder begins the process of generation. 
\paragraph{Multimodal MT.}
We extend unimodal MT with multimodal attention~\cite{calixto-etal-2016-dcu,caglayan-etal-2016-multimodality} in the decoder, in order to incorporate visual information into the baseline NMT. Let us denote the visual counterpart of textual hidden states $H$ by $V$. Multimodal attention simply applies another attention layer on top of $V$, which yields a visual context vector $c_t^{\mathbf{V}}$ at each decoding timestep $t$. The final multimodal context vector that would be given as input to the second GRU is simply the sum of both context vectors.

\paragraph{Unimodal wait-$k$ NMT.}
\label{sec:method_sim_waitk}

We explore deterministic wait-$k$~\cite{ma-etal-2019-stacl} approach as a unimodal baseline\footnote{These baselines are equivalent to the deterministic approaches used in \citet{caglayan2020simultaneous}.} for simultaneous NMT. The wait-$k$ model starts by reading $k$ source tokens and writes the first target token. The model then reads and writes one token at a time to complete the translation process. This implies that the attention layer will now attend to a partial textual representation corresponding to $k$-words. We use the decoding-only variant which does not require re-training an NMT model \ie it re-uses the already trained consecutive NMT baselines.

\subsection{Policy Learning Framework}
\label{sec:method_sim_rl}
\paragraph{RL baseline.}
We closely follow~\citet{gu-etal-2017-learning} and cast SiMT as a task of producing a sequence of \T{READ} or \T{WRITE} actions. We then devise an RL model that connects the MT system and these actions. The model is based on a reward function that takes into account both quality and latency. Following standard RL, the framework is composed of an environment and an agent. The agent takes the decision of either reading one more input token or writing a token into the output -- hence two actions are possible: \T{READ} and \T{WRITE}. The environment is a pre-trained NMT system which is \I{frozen} during RL training. 

The agent is a GRU that parameterises a stochastic policy which decides on the action $a_t$ by receiving as input the observation $o_t$.\footnote{\red{We note that the use of GRU cells is not critical for the multimodal components. They were chosen as they led to the best performance in our implementation.}} In our setup, $o_t$ is defined as $[c_t^{\mathbf{T}}; y_t; a_{t-1}]$, \ie the concatenation of vectors coming from the environment, as well as the previously produced action sequence. At each time step, the agent receives a reward $r_t = r_t^Q + r_t^D$ where $r_t^Q$ is the quality reward (the difference of smoothed BLEU scores for partial hypotheses produced from one step to another) and $r_t^D$ is the latency reward formulated as:
\begin{align*}
    r_t^{D} = \alpha \left[\text{sgn}(C_t - C^*)+1\right] + \beta \lfloor D_t - D^* \rfloor_{+}
    \vspace{-3pt}
\end{align*}
\noindent where $C_t$ denotes the consecutive wait (CW) metric which is added to avoid long consecutive waits~\cite{gu-etal-2017-learning}. CW measures how many source tokens are consecutively read between committing two translations. $D_t$ refers to average proportion (AVP)~\cite{cho2016can}, which defines the average proportion of wait tokens when translating the words. $D^*$ and $C^*$ are hyper-parameters that determine the expected/target values. The optimal quality-latency trade-off is achieved by balancing the two reward terms. In our reward implementation we again closely follow~\citet{gu-etal-2017-learning}.

\begin{figure*}
\subfloat{(a)}{\includegraphics[width=0.47\textwidth]{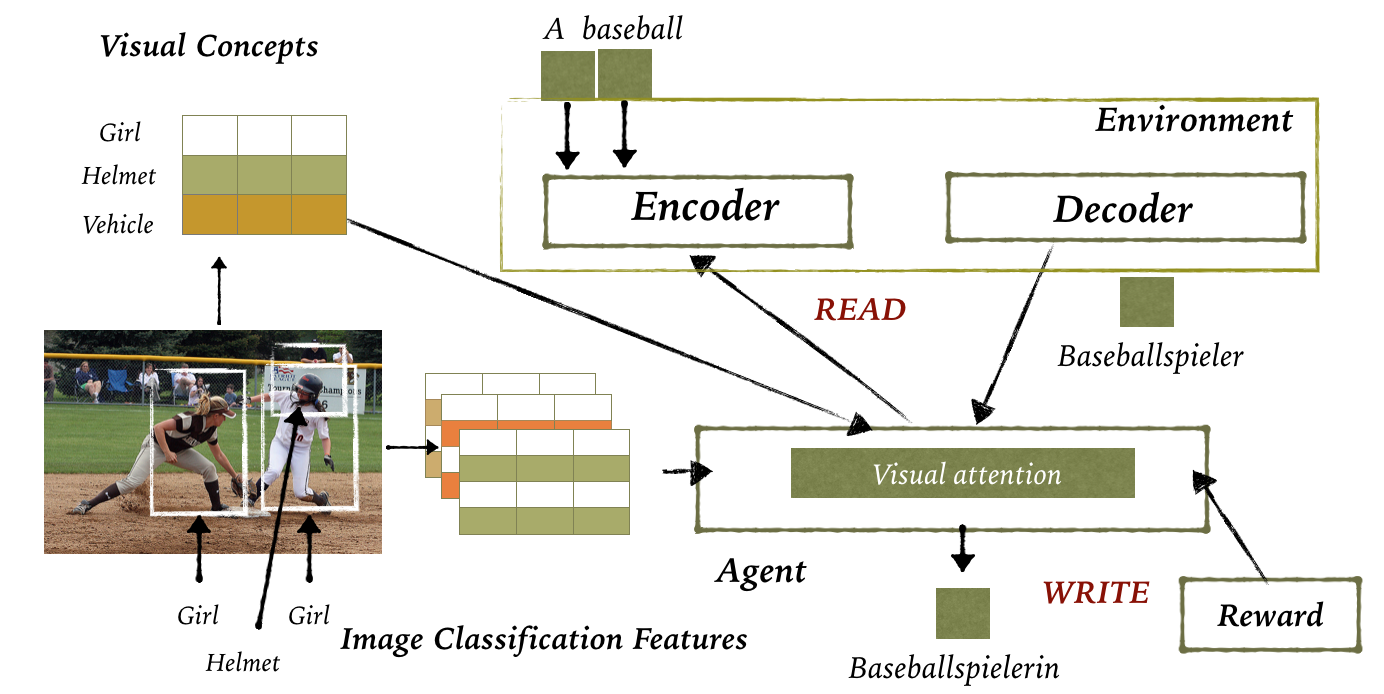}}
\subfloat{(b)}{\includegraphics[width=0.48\textwidth]{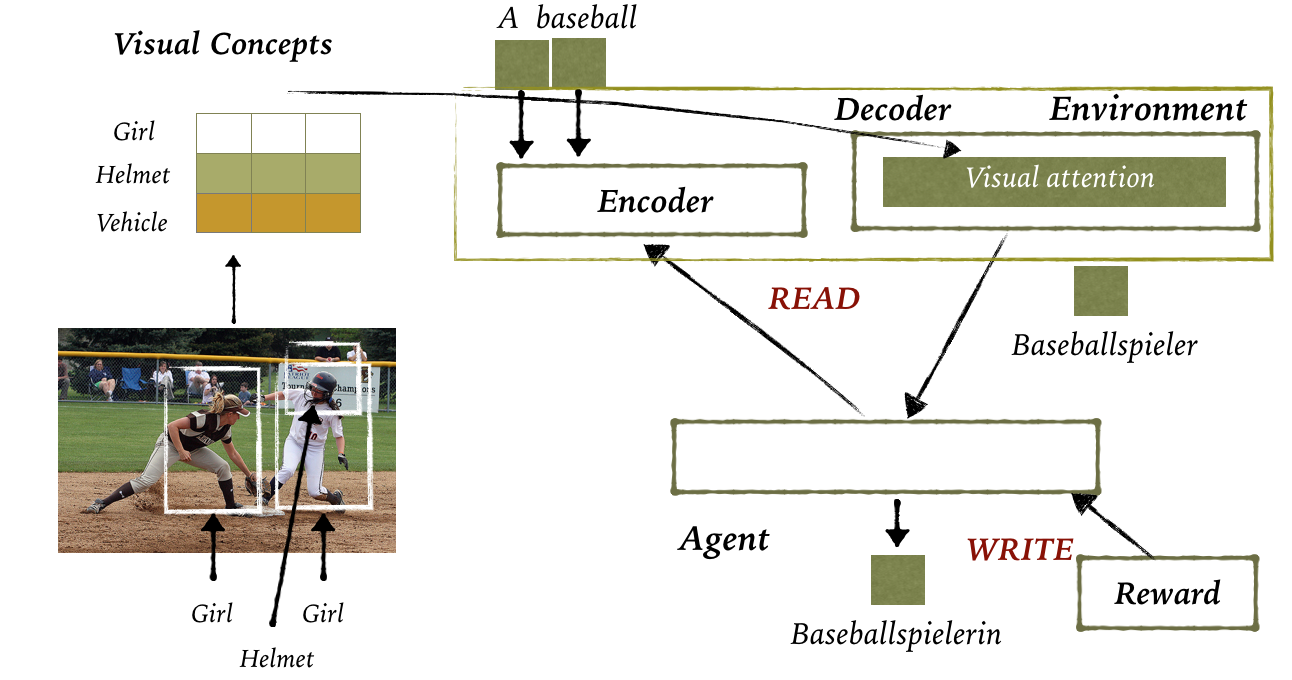}}
\caption{Our multimodal RL SiMT models: the agent interacts with the environment to receive new translation and at each time step produces the \texttt{READ/WRITE} action. For each action it receives a reward. The image information can be integrated into the agent by means of an attention mechanism (a, \att), or into the environment decoder (b, \env) producing the next translation.}
\label{fig:agent-env-att}
\end{figure*}

\paragraph{Multimodal extension.} Here we focus on integrating the visual information with the agent (see Figure~\ref{fig:agent-env-att}). The basic premise is that the addition of multimodal information, especially in the context of MMT, can result in the agent learning better and more flexible policies. We explore several ways to integrate visual information into this framework:
\begin{itemize}[leftmargin=0cm,itemindent=.5cm,labelsep=0cm,align=left]
\item \textbf{Multimodal initialisation} (\init) - the agent network is initialised with the image vector $V$ as $d_0$. We expect this vector to give the agent some context w.r.t.\ the source sentence so it can potentially read fewer words before producing outputs.
\item \textbf{Multimodal attention} (\att, Figure~\ref{fig:agent-env-att}) applies another attention layer on top of $V$, which yields a visual context vector $c_t^{\mathbf{V}}$ at each agent time step $t$. This visual context vector is a dot product attention $c_t^{\mathbf{V}} = \text{Attention}(V, \text{query}\leftarrow y_t)$ that computes the similarity between $V$ and the embedding of the target word produced by the decoder at the time step $t$. In this setting, we expect the agent to pay attention to the information in $V$ that will help in defining whether $y_t$ is good enough to be written to the output (potentially with closer relationship to some part of the image information) or we need to read more source words to produce a better $y_t$. We  concatenate $c_t^{\mathbf{V}}$ to  $o_t$, which now becomes $[c_t^{\mathbf{T}}; y_t; a_{t-1}; c_t^{\mathbf{V}}]$;

\item As a \textbf{control}, we also study \textbf{multimodal environment} (\env, Figure~\ref{fig:agent-env-att}) where we use the MMT baseline as environment. Here, we expect the initial translation quality of SiMT RL models be closer to the quality of the respective consecutive multimodal baseline as the image information is expected to compensate for partial source information. When combined with \init and \att settings, we expect the agent to exploit different kinds of image information than the environment.  
\end{itemize}

\paragraph{Learning.}
To learn the multimodal agent, we introduce an additional neural network with the same structure as that of the agent GRU network to provide for control variates (baselines) that improve the Monte-Carlo policy gradient (REINFORCE \cite{williams1992simple}). Note that here we depart from the previous work where~\citet{gu-etal-2017-learning} use a simple multilayer perceptron as the baseline.

Therefore, with the reward $r_t$ at each time step, we obtain the estimation of the gradients by subtracting the baselines $b(o_t)$:
\begin{align*}
    \nabla_{\theta}J(\theta)=\mathbb{E}[\sum_{t=0}^{T-1} \nabla_{\theta} \log \pi(a_t|o_t)(r_t - b(o_t))]
\end{align*}
To further reduce the variance of the gradient estimator, we also introduce a temperature $\tau$ for controlling the interpolation between discrete action samples and continuous categorical densities, which yields to a Gumbel-Softmax reparameterisation~\cite{jang2017categorical} that smooths the learning. \red{To be more precise, we use the Gumbel-Softmax distribution instead of argmax while sampling. So the probability of the \T{WRITE} action is given to the agent network instead of the index of the action.}

\subsection{Visual Features}\label{sec:visfeatures}
In order to represent the visual information, we explore two settings that differ in the organisation of the spatial structure. Regardless of the setting, the image features are linearly projected into the hidden space of the decoder to yield the tensor $V$.

\paragraph{Image classification features (OC)} are \I{global} image information represented by convolutional feature maps, which are believed to capture spatial cues. These features are extracted from the final convolution layer of a ResNet-50 convolutional neural network (CNN)~\cite{he2016resnet} pre-trained on ImageNet~\cite{deng2009imagenet} for object classification. The size of the final feature tensor being 8x8x2048, the visual attention is applied on a grid of 64 equally-sized regions. 

\paragraph{Visual Concepts (VC)} are explicit object representations where \I{local} regions are detected as objects and subsequently encoded with 100-dimensional word representations. For a given image, the detector provides 36 object and 36 attribute region proposals which are abstract concepts associated with the image. We represent each of the detected region with its corresponding GloVe~\cite{pennington-etal-2014-glove} word vectors. An image is thus represented by a feature tensor of size 72x100 and the visual attention is now applied on these visual concepts, rather than the uniform grid of the first approach above. We hypothesise that this type of information can result in better referential grounding by using conceptually meaningful units rather than global features.
The detector used here is a Faster R-CNN/ResNet-101 object detector (with 1600 object labels)~\cite{butd}\footnote{https://hub.docker.com/r/airsplay/bottom-up-attention} pre-trained on the Visual Genome dataset~\cite{visualgenome}.

%% file: setup.tex
\subsection{Dataset}
\label{sec:setup_dataset}
We perform experiments on the Multi30k dataset~\cite{elliott-etal-2016-multi30k}\footnote{https://github.com/multi30k/dataset} which extends the Flickr30k image captioning dataset~\cite{young-etal-2014-image} with caption translations in German and French~\cite{elliott-etal-2017-findings}.
Multi30k is a standard MMT dataset that contains parallel sentences in two languages that describe the images. The training set for each language direction comprises 29,000 image-source-target triplets whereas the development and the test sets have around 1,000 samples. We use the corresponding test sets from 2016, 2017 and 2018 for evaluation. 
\paragraph{Pre-processing.} We use Moses scripts~\cite{koehn-etal-2007-moses} to lowercase, normalise and tokenise the sentences. We then create word vocabularies on the \I{training} subset of the dataset. We did not use subword segmentation to avoid its potential side effects on fixed policy SiMT and to be able to better analyse the grounding capability of the models. The resulting English, French and German vocabularies contain 9.8K, 11K and 18K tokens, respectively.  

\subsection{Evaluation}
\label{sec:setup_eval}
We use {\bf BLEU}~\cite{papineni-etal-2002-bleu} for quality, and perform significance testing via bootstrap resampling using the \texttt{Multeval} tool \cite{clark-etal-2011-better}.
For latency, we measure \B{Average proportion} (AVP)~\cite{cho2016can}. AVP is the average number of source tokens required to commit a translation. This metric is sensitive to the difference in lengths between source and target. Hence, as our main latency metric we measure  \B{Average Lagging} (AVL)~\cite{ma-etal-2019-stacl} which estimates the number of tokens the ``writer'' is lagging behind the ``reader'', as a function of the number of input tokens read.

\subsection{Training}
\label{sec:setup_training}
\paragraph{Hyperparameters.}
We set the embeddings dimensionality and GRU hidden states to 200 and 320, respectively. 
We use the ADAM~\cite{kingma2014adam} optimiser with the learning rate 0.0004 and the batch size of 64. 
We use \T{pysimt}~\cite{caglayan2020simultaneous} with PyTorch~\cite{paszke2017automatic} v1.4 for our experiments.\footnote{https://github.com/ImperialNLP/pysimt} We early stop w.r.t.\ the validation BLEU with the patience of 10 epochs. 
On a single NVIDIA RTX2080-Ti GPU, the training takes around 35 minutes for the unimodal model and around 1 hour for the multimodal model. The number of learnable parameters is between 6.9M and 9.3M depending on the language pair and the type of multimodality.

For the \B{RL systems}, we  follow~\cite{gu-etal-2017-learning}.\footnote{https://github.com/nyu-dl/dl4mt-simul-trans} The agent is implemented by a 320-dimensional GRU followed by a softmax layer and the baseline network is similar to the agent except with a scalar output layer.\footnote{Note that that~\citet{gu-etal-2017-learning} use a 2-hidden layer feed-forward network as the baseline network. In our implementation GRUs have demonstrated better performance.} We use ADAM as the optimiser and set the learning rate and mini-batch size to 0.0004 and 6, respectively. For each sentence pair in a batch, 5 trajectories are sampled. Following best practises in RL, the baseline network is trained to reduce the MSE loss between the predictions and the rewards using a second optimiser. 

For inference, greedy sampling is used to pick action sequences. We set the hyperparameters $C^*\mathrm{=}2$, $D^*\mathrm{=}0.3$, $\alpha\mathrm{=}0.025$ and $\beta\mathrm{=}-1$. To encourage exploration, the negative entropy policy term is weighed empirically with 0.001. Following~\cite{gu-etal-2017-learning}, we choose the model that maximises the quality-to-latency ratio (BLEU/AVP) on the validation set with a patience of 5 epochs.\footnote{We also attempted to choose the model that maximises BLEU or BLEU/AVL but those stopping criteria resulted in instability of convergence.} On a single NVIDIA RTX2080-Ti GPU, the training takes around 2 hours. The number of learnable parameters is around 6M.

\paragraph{Model configurations.}
We experiment with seven different configurations (below). We consider visual concepts (VC) as the main source of multimodal information. Visual concepts are more abstract forms of multimodal information. Unlike spatial image representation or region of interest-based object representations, where the representation for the same concept can vary significantly across images, visual concepts remain constant. For example, the visual concept ``dog'' is the same regardless of the breed, colour, size, position, etc. of the concept in different images. Image classification (OC) features are used as a contrastive setting. 
 
\begin{itemize}
\item Unimodal RL baseline (\base): This baseline  follows~\cite{gu-etal-2017-learning} where the environment is a text-only NMT model.  
\item Multimodal agent with VC initialisation (\init VC): We initialise the agent GRU using a projection of the flattened 72x100 matrix of visual concepts.
\item Multimodal agent with attention over VC (\att VC): The agent attends over the set of visual concepts at each step.
\item  Multimodal agent with attention over OC (\att OC): The agent attends over the set of image classification-based spatial feature maps at each step.
\item Visually initialised multimodal agent with attention over VC (\initatt VC): Similar to \att VC but the agent is also initialised with VC. 
\item Multimodal environment with unimodal RL agent (\env VC): The environment is an MMT model, however the agent is a standard RL agent akin to the baseline. 
\item Multimodal agent with multimodal environment (\envinitatt VC): This merges all the variants in that both the multimodal environment and the multimodal agent attend to visual concepts, the latter is also initialised with visual information.
\end{itemize}

%% file: results.tex
\begin{table*}[t!]
\centering
\scalebox{0.8}{
\begin{tabular}{lllcc@{\hspace{1cm}}lcc@{\hspace{1cm}}lcc}
\toprule
& & \multicolumn{3}{c}{\textbf{test 2016}\phantom{spa}} & \multicolumn{3}{c}{\textbf{test 2017}\phantom{spa}} & \multicolumn{3}{c}{\textbf{test 2018}} \\
& & BLEU$\uparrow$ &  AVL$\downarrow$ &  AVP$\downarrow$ & BLEU$\uparrow$ &  AVL$\downarrow$ &  AVP$\downarrow$ &  BLEU$\uparrow$ &  AVL$\downarrow$ &  AVP$\downarrow$ \\  \midrule
\multirow{13}{*}{\rotatebox{90}{\textbf{English~$\longrightarrow$~French}}}
& Consecutive & 58.0 & 13.1 & 1.0 & 50.6 & 11.1 & 1.0 & 36.0 & 13.8 & 1.0 \\
& ~~~+VC & 59.1 & 13.1 & 1.0 & 51.0 & 11.1 & 1.0 & 36.5 & 13.8 & 1.0 \\
\cline{2-11}
&  Wait-2 & 48.1 & 2.6 & 0.7 & 42.9 & 2.6 & 0.7 & 32.1 & 2.7 & 0.7 \\
&  Wait-3 & 54.0 & 3.5 & 0.7 & 48.6 & 3.5 & 0.7 & 35.5 & 3.5 & 0.7\\
\cline{2-11}
& \tt{RL} & 50.8 & 3.3 & \bf 0.7 & 44.3 & 3.0 & \bf 0.7 & 32.1 & 3.5 & \bf 0.7 \\
&  ~~~+{\tt att}-OC  & 53.0* & 4.1 & 0.8 & 46.4* & 3.9 & 0.8 & 33.3* & 4.4 & 0.8 \\
& ~~~+{\tt att}-VC & 53.0* & 4.0 & \bf 0.7  & 46.5* & 3.7 & 0.8 & 33.3* & 4.2 & \bf 0.7 \\
& ~~~+{\tt init}-VC & 49.6 & \bf 2.8 & \bf 0.7 & 43.3 & \bf 2.6 & \bf 0.7 & 31.5 & \bf 2.9 & \bf 0.7\\
& ~~~+{\tt init-att}-VC & 52.6* & 3.8 & \bf 0.7 & 46.3* & 3.6 & \bf 0.7 & 33.3* & 4.1 & \bf 0.7 \\
& ~~~+{\tt env}-VC & 54.0* & 3.3 & \bf 0.7 & 47.2* & 3.1 & \bf 0.7 & 33.7* & 3.4 & \bf 0.7 \\
& ~~~+{\tt env-init-att}-VC & \bf 54.0* & 3.9 & \bf 0.7 & \bf 47.7* & 3.8 & 0.8 & \bf 34.4* & 4.2 & \bf 0.7\\
\midrule\midrule
\multirow{13}{*}{\rotatebox{90}{\textbf{English~$\longrightarrow$~German}}}
& Consecutive & 35.5 & 13.1 & 1.0 & 27.7 & 11.1 & 1.0 & 25.8 & 13.8 & 1.0 \\
& ~~~+VC & 35.9 & 13.1 & 1.0 & 27.0 & 11.1 & 1.0 & 25.4 & 13.8 & 1.0 \\
\cline{2-11}
& Wait-2 & 28.3 & 2.2 & 0.6 & 22.5 & 2.2 & 0.7 & 20.1 & 2.2 & 0.6\\
& Wait-3 & 32.6 & 3.0 & 0.7 & 25.4 & 3.0 & 0.7 & 24.1 & 3.0 & 0.7\\
\cline{2-11}
& \tt{RL} & 31.0 & 2.7 & 0.7 & 23.0 & 2.6 & 0.7 & 22.0 & 2.7 & 0.7\\
&  ~~~+{\tt att}-OC & 33.9* & 3.7 & 0.7 & \bf 25.8* & 3.4 & 0.7 & \bf 24.5* & 3.8 & 0.7\\
& ~~~+{\tt att}-VC & 33.3* & 3.3 & 0.7 & 24.7* & 3.0 & 0.7 & 23.0* & 3.2 & 0.7\\
& ~~~+{\tt init}-VC & 29.7 & 2.8 & 0.7 & 21.3 & 2.4 & 0.7 & 20.5 & 2.5 & 0.6\\
& ~~~+{\tt init-att}-VC & \bf 34.1* & 3.3 & 0.7 & 25.3* & 3.1 & 0.7 & 24.1* & 3.4 & 0.7 \\
& ~~~+{\tt env}-VC & 30.0 & \bf 2.5 & \bf 0.6 & 21.7 & \bf 2.2 & \bf 0.6 & 19.7 & \bf 2.2 & \bf 0.6 \\
& ~~~+{\tt env-init-att}-VC & 31.4 & 3.0 & 0.7 & 24.0* & 2.9 & 0.7 & 22.4 & 3.0 & 0.7 \\
\bottomrule
\end{tabular}%
}
\caption{\label{table:main_res} Results for the test sets 2016, 2017 and 2018 (averaged over 3 runs): * marks statistically significant increases in BLEU w.r.t.\ \base (p-value $\leq 0.05$). Bold highlights best scores across the RL approaches.}
\end{table*}

In this section, we first provide the results from our experiments (Section \ref{sec:quantres}) and then analyse the behaviour of the (multimodal) agents (Section \ref{sec:att_analysis}). 

\subsection{Quantitative Results}\label{sec:quantres}
\paragraph{SiMT vs.\ Consecutive.} We present the main results in Table~\ref{table:main_res}. The top block for each language pair shows the textual Consecutive model and its multimodal counterpart (Consecutive+VC). These are our upperbounds since they have access to the entire source before translating. As expected, they have better BLEU but much larger AVL. 

\paragraph{RL SiMT vs.\ Deterministic policy.} The second block in Table~\ref{table:main_res} shows the deterministic policy Wait-$2$ and Wait-$3$ approaches. \base performs on par with the Wait-$2$ (English-French) and Wait-$3$ (English-German).   
We however emphasise the flexibility of the stochastic policies with RL models. These are particularly beneficial in the multimodal scenario and allow for exploitation of the image information more efficiently especially towards reducing the average lag. We further expand on this later in Section~\ref{sec:att_analysis}.

\paragraph{Unimodal RL vs.\ Multimodal RL.} The third block in Table~\ref{table:main_res} compares all multimodal RL variants against the text-only SiMT RL (\base). In general, the multimodal RL models produce translations that are significantly better than \base. 
\paragraph{Across Multimodal RL Setups.} With regard to different configurations, we observe (1) an increase in quality for the \att models when compared to \base which is consistent in both types of visual inputs OC and VC, and (2) a decrease in the lag for the \init models at a small decrease in quality  (for VC \init in comparison to \base). 

This observation suggests that the RL model with the agent explicitly attending over image information leads to an increase in quality, as the multimodal agent model is more selective towards the word choice. 
The \init configuration with prior image context on the other hand reduces the lag and seems to use \T{WRITE} actions more often than \T{READ} actions. It is interesting that OC and VC features result in similar quality translations, however we see that on average the average lag is lower with VC. We hypothesise that this could be due to the fact that the representations remain constant across images (see Section~\ref{sec:setup_training}).

The \initatt configuration represents a middle ground and we see similar quality improvement to \att across setups (a gain of 2 BLEU points on average) but with a slightly lower latency. 
We however observe that \envinitatt has a slightly inferior performance with a a pronounced latency when compared to the \env model. We investigate this aspect in the next sections. 

\begin{figure*}[t]
\centering
\includegraphics[width=0.6\textwidth]{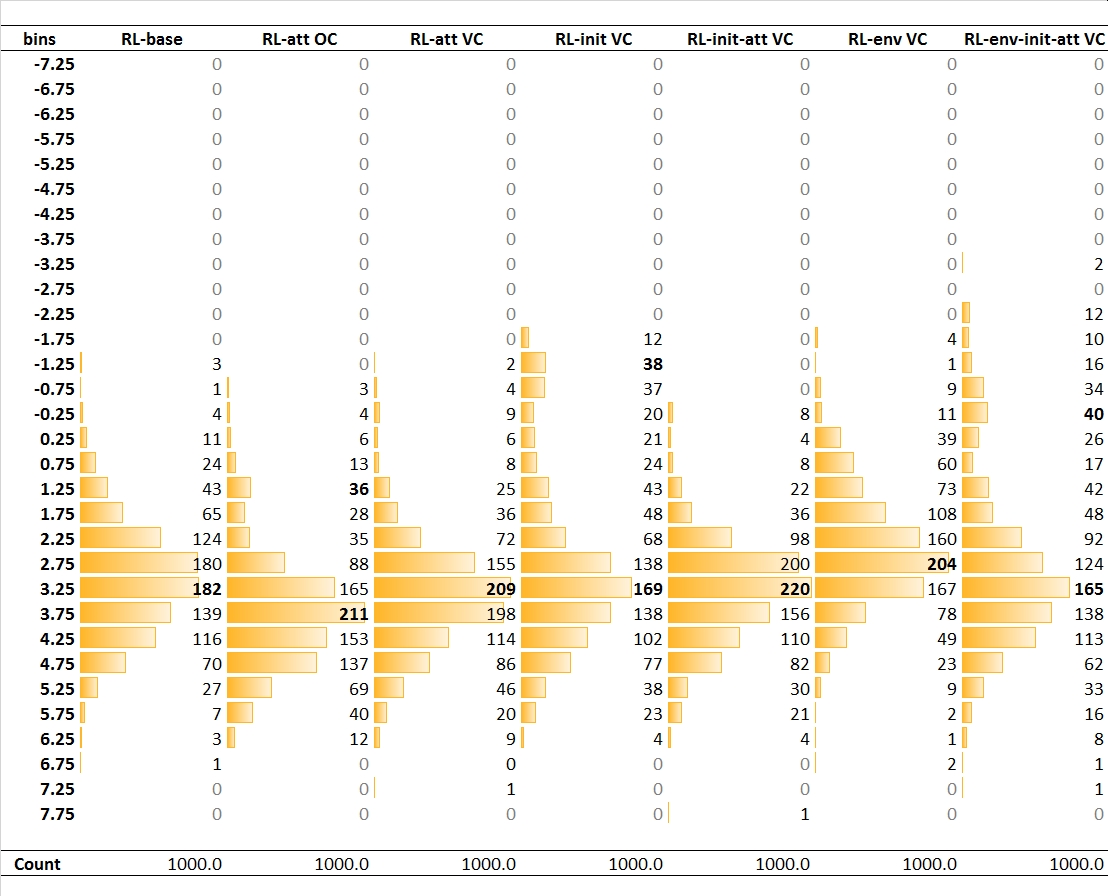}
\caption{Histogram of per sentence lag values in test 2016 English-German. Y axis shows mean values per bin. Bold highlights modes for each distribution.}
\label{fig:lag-bins}
\end{figure*}

\begin{figure*}[h]
\centering
\includegraphics[width=0.6\textwidth]{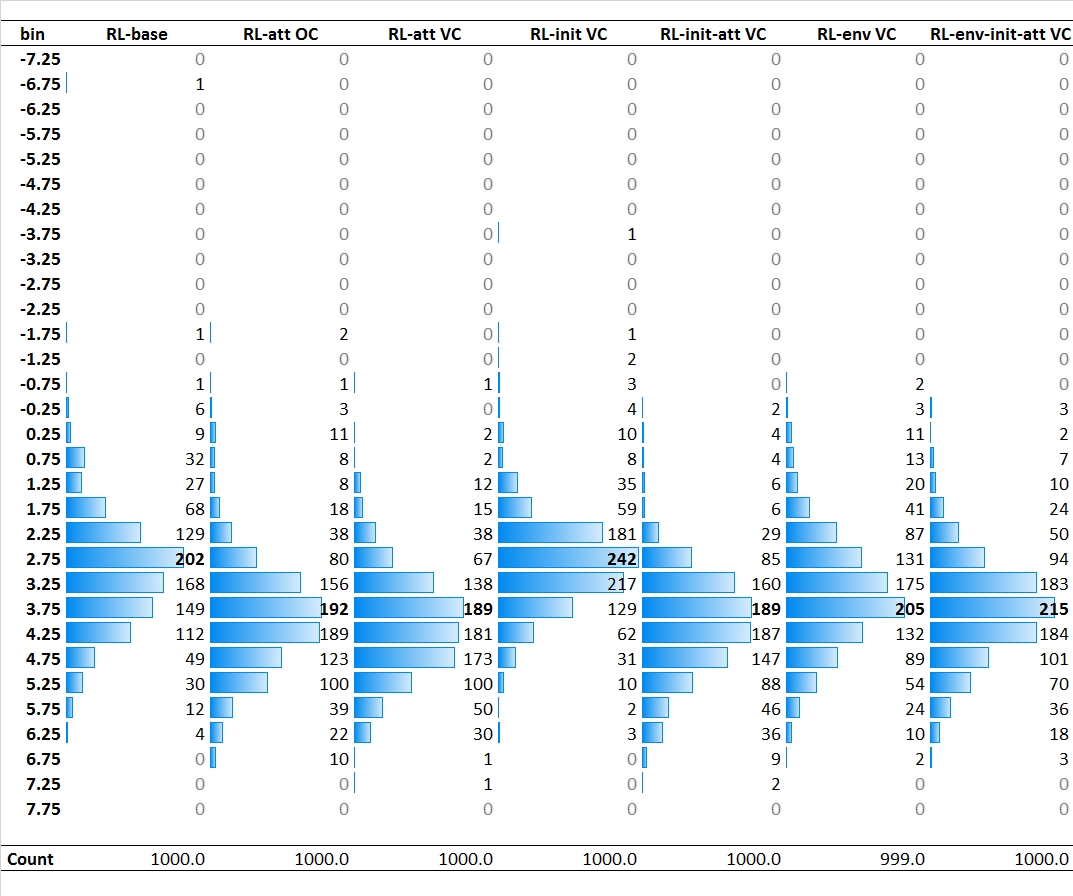}
\caption{Histogram of per sentence lag values for test 2016 English-French. Y axis shows mean values per bin. Bold highlights modes for each distribution.}
\label{fig:en-fr-att-norms}
\end{figure*}

\paragraph{Investigating Average Lag.} To further study the impact of our configurations on the sentence level lag, in Figure~\ref{fig:lag-bins} we present the binned-histograms of sentence lags over the English$\rightarrow$German test 2016 set. Generally, the models which are initialised with image information seem to have more mass towards the smaller delay bins.  
In terms of \init and \envinitatt setups, we also observe the presence of two modes around the lag value 3 as well as around two negative values (around -0.25 and -1.25 respectively). These negative lag values are due the difference in length between source and target sentences which is typical for the English$\rightarrow$German. This also shows that the agent initialised with the image information tends to prefer \T{WRITE} actions with fewer \T{READ} actions. Further, on manual inspection of some samples, we observed that in the cases with negative lag the model begins with a \T{WRITE} action straight after reading the first token (See Table~\ref{table:neg-lag-example}). As the agent is a GRU model, this behavior resembles that of an image captioning model. We also observe similar trends for English$\rightarrow$French with \init models predominantly having more mass towards smaller delay bins (see Figure~\ref{fig:en-fr-att-norms}).

\begin{figure}[t]
\centering
\includegraphics[width=0.45\textwidth]{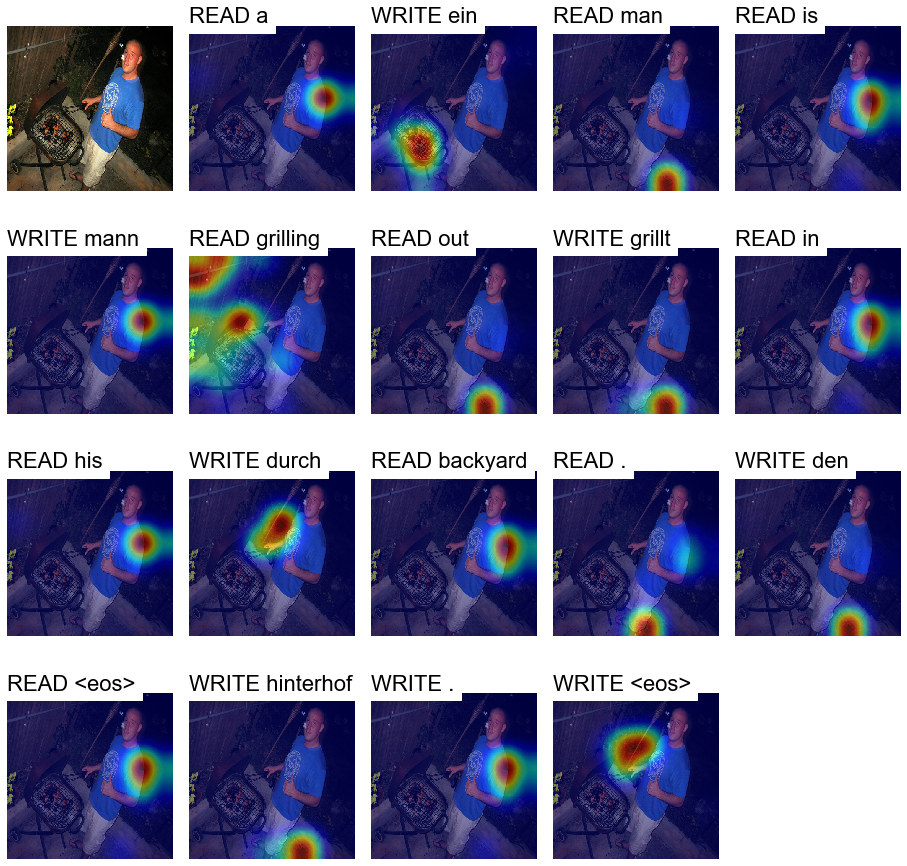}
\caption{Visualisation of the agent attention and the corresponding actions over the source sentence from the test2016: `A man is grilling out in his backyard.'}
\label{fig:agent-resnet-att}
\end{figure}

\begin{table*}[h]
\centering
\scalebox{0.9}{
\begin{tabular}{lr}
\toprule
\T{SRC:} the red car is ahead of the two cars in the background . & \\
\T{REF:} das rote auto fährt vor den beiden autos im hintergrund .\\
`the red car goes before the both cars in the background' & \\
\T{\init:} die person ist im begriff , die rote mannschaft auf dem roten auto versammelt . \\
`the person is in concept, that red manhood on the red car gathered'& \\
\T{Actions:} 0 1 1 1 1 1 1 1 1 1 1 1 1 1 1 1 1 1 1 1 1 1 1 0 0 1 0 0 1 0 0 1  & \\
\T{BLEU:} 3.7  & \\
\T{LAG:} -1.875  & \\
\bottomrule
\end{tabular}}
\caption{Example of a German VC \init setup sentence with a negative lag, where the model tends to write more before reading  new words.}
\label{table:neg-lag-example}
\end{table*}

\subsection{Agent Attention over Visual Inputs}\label{sec:att_analysis}

In Figure~\ref{fig:agent-resnet-att} we visualize the agent's attention at each time step. On average, the agent actions correlate with the objects it attends to when producing the translation. 

\begin{figure}
\centering
\includegraphics[width=0.48\textwidth]{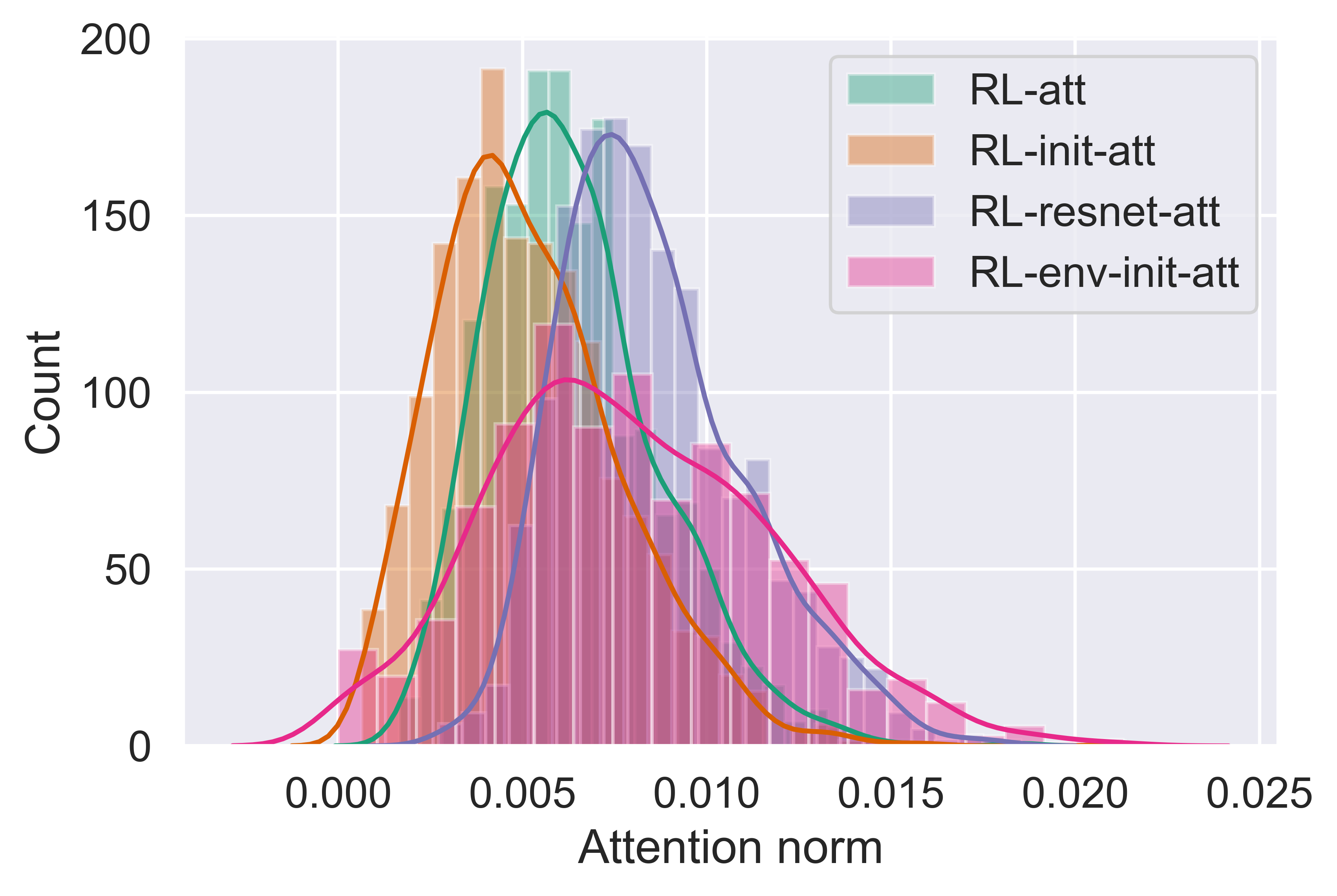}
\caption{Distribution of attention norms for different agents with visual attention trained on the English$\rightarrow$German dataset.}
\label{fig:en-de-att-norms}
\end{figure}

We now examine the general pattern of agent attention over the visual concepts across the four configurations using attention norm: a) \att-VC; b) \att-OC; c) \initatt; and d) \envinitatt. The attention norm is simply the average $\ell_2$ norm between two consecutive attention time-steps. This can help in measuring the average visual attention per time step for a given sentence. We then compare the attention norm distributions over all the sentences in the English$\rightarrow$German test 2016 set for the four different agent attention configurations. We present the result in Figure~\ref{fig:en-de-att-norms}. Overall, \init and \att models are significantly more peaky than the \envinitatt. This suggests that \envinitatt model is generally spread across the 72 visual concepts more uniformly than the other two models. This perhaps is one of the causes for the slightly inferior performance of the model. 
We hypothesise that further regularisation of the attention distribution can ameliorate this behavior and leave it as future work. 

%% file: conclusion.tex
In this paper we presented the first thorough exposition of multimodal reinforcement learning strategies for simultaneous machine translation. We demonstrate the efficacy of visual information and show that it leads to adaptive policies which substantially improve over the deterministic and unimodal RL baselines. Our empirical results indicate that both agent-side and environment-side visual information can be exploited to achieve higher quality translations with lower latency. 

Throughout the experimental journey, we observed that the optimisation of simultaneous machine translation for dynamic policies is non-trivial, due to the two competing objectives: translation quality versus latency. 
For unimodal simultaneous machine translation, RL approaches tend to achieve translation quality on par with the quality of the deterministic policies within the same average lag. We believe that the fundamental issue is related to the high variance of the estimator for sequence prediction, which increases sample complexity and impedes effective learning. On the other hand, the approaches with deterministic policies are simple and effective, as they are positively biased for language pairs that are close to each other. But the latter suffer from poor generalisation.

In the multimodal simultaneous machine translation setting, however, the variance of the estimator from RL models can be substantially reduced with to the presence of additional (visual) information.